\documentclass{article}
\usepackage[numbers]{natbib}
\usepackage{graphicx}
\usepackage{wrapfig}
\usepackage{jmlr2e}
\usepackage{times}

\usepackage{tikz}
\usepackage[hidelinks]{hyperref}
\usepackage{comment}
\usepackage{amsmath,amssymb}
\usepackage{color}
\usepackage{booktabs, bbm, multirow, makecell, lipsum, tabularx}
\usepackage{caption}
\usepackage[ruled,vlined]{algorithm2e}
\newcolumntype{Y}{>{\centering\arraybackslash}X}

\begin{document}
\newcommand*\samethanks[1][\value{footnote}]{\footnotemark[#1]}
\title{Subspace Diffusion Generative Models}
\author{Bowen Jing,\thanks{Equal contribution} \; Gabriele Corso,\samethanks \; Renato Berlinghieri, Tommi Jaakkola \\
       {\texttt{\{bjing, gcorso, renb\}@mit.edu, tommi@csail.mit.edu}}\\
       Massachusetts Institute of Technology \\
}

\maketitle
\thispagestyle{empty}

\begin{abstract}
Score-based models generate  samples by mapping noise to data (and vice versa) via a high-dimensional diffusion process. We question whether it is necessary to run this entire process at high dimensionality and incur all the inconveniences thereof. Instead, we restrict the diffusion via projections onto \emph{subspaces} as the data distribution evolves toward noise. When applied to state-of-the-art models, our framework simultaneously \emph{improves} sample quality---reaching an FID of 2.17 on unconditional CIFAR-10---and \emph{reduces} the computational cost of inference for the same number of denoising steps. Our framework is fully compatible with continuous-time diffusion and retains its flexible capabilities, including exact log-likelihoods and controllable generation. Code is available at \url{https://github.com/bjing2016/subspace-diffusion}.
\end{abstract}

\section{Introduction} \label{sec:intro}

\begin{figure}[t]
    \centering
    \includegraphics[width=0.9\textwidth]{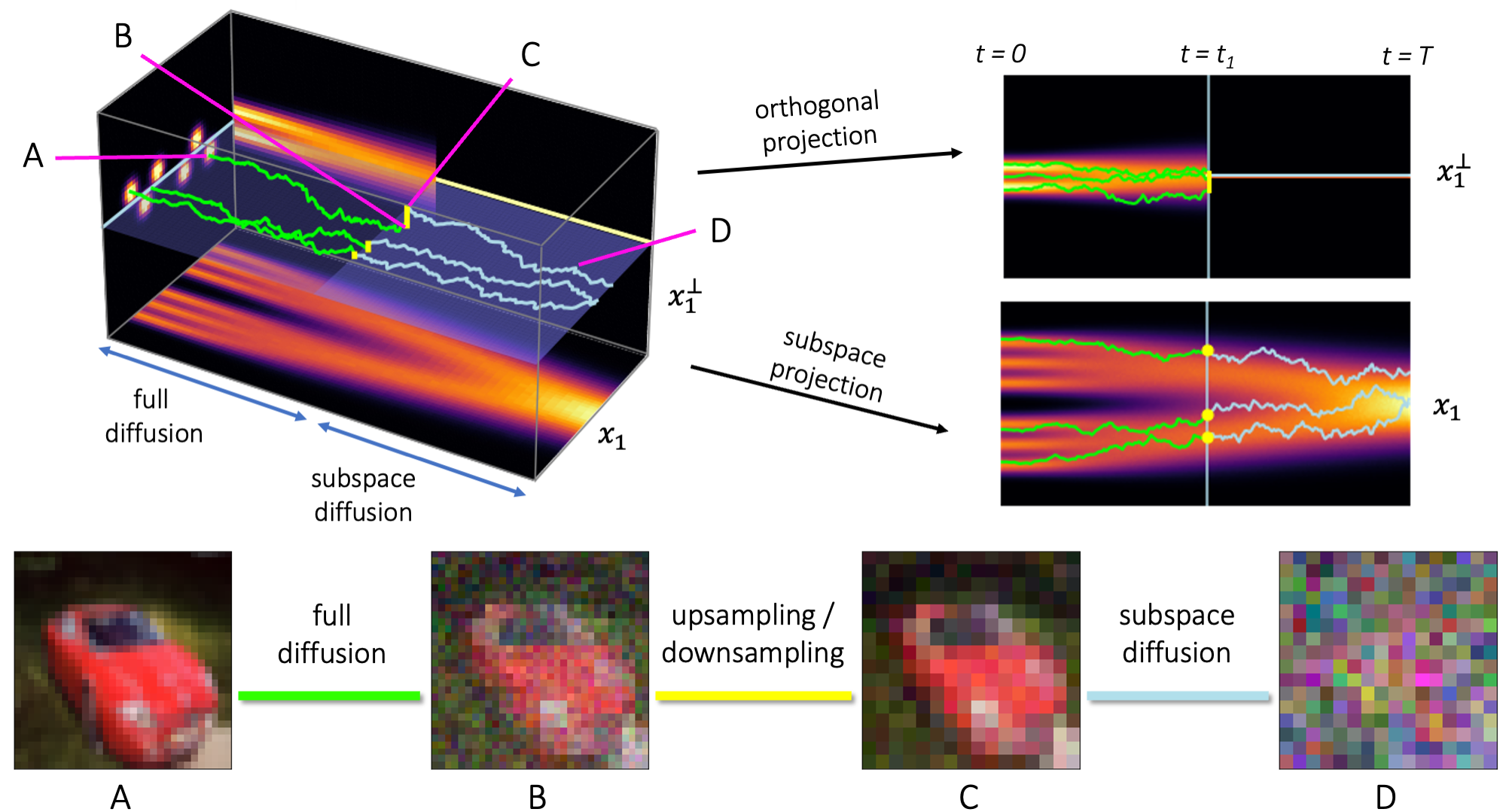}
    \caption{Visual schematic of subspace diffusion with one projection step. \emph{Top left}: The starting data distribution $\mathbf{x}_0(0)$ lies near a subspace (light blue line). As the data evolves, the distribution of the orthogonal component $\mathbf{x}_1^\perp(t)$ approaches a Gaussian faster than the subspace component $\mathbf{x}_1(t)$. At time $t_1$ we project onto the subspace and restrict the remaining diffusion to the subspace. To generate data, we use the full and subspace score models to reverse the full and subspace diffusion steps, and sample $\mathbf{x}_1^\perp(t_1)$ from a Gaussian to reverse the projection step. \emph{Top right}: The diffusion of the subspace component $\mathbf{x}_1(t)$ is unaffected by the projection step and restriction to the subspace; while the orthogonal component is diffused until $t_1$ and discarded afterwards. \emph{Bottom}: CIFAR-10 images corresponding to points along the trajectory, where the subspaces correspond to lower-resolution images and projection is equivalent to downsampling.}
    \label{fig:method}
\end{figure}

Score-based models are a class of generative models that learn the score of the data distribution as it evolves under a diffusion process in order to generate data via the reverse process \cite{song2021score,ho2020denoising}. These models---also known as diffusion models---can generate high-quality and diverse samples, evaluate exact log-likelihoods, and are easily adapted to conditional and controlled generation tasks \cite{song2021score}. On the CIFAR-10 image dataset, they have recently achieved state-of-the-art performance in sample generation and likelihood evaluation  \cite{vahdat2021score,kingma2021variational}.

Despite these strengths, in this work we focus on and aim to address a drawback in the current formulation of score-based models: the forward diffusion occurs in the full ambient space of the data distribution, destroying its structure but retaining its high dimensionality. However, it does not seem parsimonious to represent increasingly noisy latent variables---which approach zero mutual information with the original data---in a space with such high dimensionality. The practical implications of this high latent dimensionality are twofold: 

\emph{High-dimensional extrapolation}. The network must learn the score function over the entire support of the high-dimensional latent variable, even in areas very far (relative to the scale of the data) from the data manifold. Due to the curse of dimensionality, much of this support may never be visited during training, and the accuracy of the score model in these regions is called into question by the uncertain extrapolation abilities of neural networks \cite{xu2020neural}. Learning to match a lower-dimensional score function may lead to refined training coverage and further improved performance.

\emph{Computational cost}. Hundreds or even thousands of evaluations of the high-dimensional score model are required to generate an image, making inference with score-based models much slower than with GANs or VAEs \cite{ho2020denoising,song2021score}. A number of recent works aim to address this challenge by reducing the number of steps required for inference \cite{song2020denoising,salimans2021progressive,jolicoeur2021gotta,nichol2021improved,dhariwal2021diffusion,kong2021fast,watson2021learning,san2021noise,lam2021bilateral,bao2022analytic}. However, these methods generally trade off inference runtime with sample quality. Moreover, the dimensionality of the score function---and thereby the computational cost of a single score evaluation---is an independent and equally important factor to the overall runtime, but this factor has received less attention in existing works.

\textbf{Subspace diffusion models} aim to address these challenges. In many real-world domains, target data lie near a linear subspace, such that under isotropic forward diffusion, the components of the data orthogonal to the subspace become Gaussian significantly before the components in the subspace. We propose to use a full-dimensional network to model the score only at lower noise levels, when all components are sufficiently non-Gaussian. At higher noise levels, we use smaller networks to model in the subspace only those components of the score which remain non-Gaussian. As this reduces both the number and domain of queries to the full-dimensional network, subspace diffusion addresses both of our motivating concerns. Moreover, in contrast to many prior works, subspace diffusion remains fully compatible with the underlying continuous diffusion framework \cite{song2021score}, and therefore preserves all the capabilities available to continuous score-based models, such as likelihood evaluation, probability flow sampling, and controllable generation. 

While subspace diffusion can be formulated in fully general terms, in this work we focus on generative modeling of natural images. Because the global structure of images is dominated by low-frequency visual components---i.e., adjacent pixels values are highly correlated---images lie close to subspaces corresponding to lower-resolution versions of the same image. Learning score models over these subspaces has the advantage of remaining compatible with the translation equivariance of convolutional neural networks, and therefore requires no architectural modifications to the score model.

\paragraph{\textbf{Contributions.}} We formulate the diffusion process, training procedure, and sampling procedure in subspaces; to our knowledge, this represents the first investigation of dimensionality reduction in a score-based model framework. We develop a method, the \emph{orthogonal Fisher divergence}, for choosing among candidate subspaces and the parameters of the subspace diffusion. Experimentally, we train and evaluate lower-dimensional subspace models in conjunction with state-of-the-art pretrained full-dimensional models from \cite{song2021score}. We improve over those models in sample quality and runtime, achieving an FID of 2.17 and a IS of 9.99 on CIFAR-10 generation. Finally, we demonstrate probability flow sampling and likelihood evaluation with subspace models.

\section{Background and Related Work} \label{sec:background}

\paragraph{\textbf{Score-based models.}} In score-based models, one considers the data distribution $\mathbf{x}(0) \in \mathbb{R}^d$ to be the starting distribution for a continuous diffusion process, defined by an Ito stochastic differential equation (SDE)
\begin{equation} \label{eq:forward}
    d\mathbf{x} = \mathbf{f}(\mathbf{x}, t)\;dt + \mathbf{G}(\mathbf{x}, t)\;d\mathbf{w} \quad t \in (0, T)
\end{equation}
known as the \emph{forward process}, which transforms $\mathbf{x}(0)$ into (approximately) a simple Gaussian $\mathbf{x}(T)$. By convention, we typically set $T=1$. A neural network is then trained to model the score $\nabla_\mathbf{x} \log p(\mathbf{x}, t)$ conditioned on $t$. Solving the reverse stochastic differential equation
\begin{equation} \label{eq:reverse}
    d\mathbf{x} = \mathbf{f}(\mathbf{x}, t)\;dt - \mathbf{G}(\mathbf{x}, t)\mathbf{G}(\mathbf{x}, t)^T\nabla_\mathbf{x} \log p(\mathbf{x}, t) \; dt + \mathbf{G}(t) \;d\mathbf{\bar{w}}   
\end{equation}
starting with samples from the simple Gaussian distribution $\mathbf{x}(T)$ yields samples from the data distribution $\mathbf{x}(0)$ \cite{song2021score,anderson1982reverse}. Score-based models were originally  formulated separately in terms of denoising score matching at multiple noise scales \cite{song2019generative}; and of reversing a discrete-time Markov chain of diffusion steps \cite{ho2020denoising}. Due to the latter formulation (associated with the term \emph{diffusion model}), $\mathbf{x}(t)$ for $t > 0$ are often referred to as \emph{latents} of $\mathbf{x}(0)$, and the simple Gaussian $\mathbf{x}$ as the \emph{prior}. The two views are unified by the observation that the variational approximation to the reverse Markov chain matches the score of the diffused data \cite{song2021score}.

The score model $s_\theta(\mathbf{x}, t)$ can be trained via denoising score matching \cite{song2019generative} using the perturbation kernels $p(\mathbf{x}(t) \mid \mathbf{x}(0))$, which are analytically determined by $\mathbf{f}(\mathbf{x}, t), \mathbf{G}(t)$ at each time $t$. The learned score can be readily adjusted with fixed terms for controlled generation tasks in the same manner as energy-based models \cite{du2019implicit}. Finally, the reverse stochastic differential equation produces the same marginals $\mathbf{x}$ as the \emph{ordinary} differential equation (ODE)
\begin{equation} \label{eq:ode}
    d\mathbf{x} = \mathbf{f}(\mathbf{x}, t)\;dt - \frac{1}{2}\mathbf{G}(\mathbf{x}, t)\mathbf{G}(\mathbf{x}, t)^T\nabla_\mathbf{x} \log p(\mathbf{x}, t) \; dt 
\end{equation}
which enables evaluation of exact log-likelihoods, but empirically results in degraded quality when used for sampling \cite{song2021score}.

\paragraph{\textbf{Accelerating score-based models.}} Due to the fine discretization required to solve \eqref{eq:reverse} to high accuracy, score-based models suffer from slow inference. Several recent works aim to address this. Denoising diffusion implicit models (DDIM) \cite{song2020denoising} can be viewed as solving the equivalent ODE with a reduced number of steps. Progressive distillation \cite{salimans2021progressive} proposes a student-teacher framework for learning sampling networks requiring logarithmically fewer steps. \cite{jolicoeur2021gotta} derives an adaptive step-size solver for the reverse SDE. Other works \cite{nichol2021improved,dhariwal2021diffusion,kong2021fast,watson2021learning,san2021noise,lam2021bilateral,bao2022analytic} focus on reducing the number of steps in the discrete-time Markov chain formulation. However, these approaches generally result in degraded sample quality compared to the best continuous-time models.

Taking a different approach, latent score-based generative models (LSGM) \cite{vahdat2021score} use a score-based model as the prior of a deep VAE, resulting in more Gaussian scores, improved sample quality, and fewer model evaluations. In a similar vein, critically-damped Langevin diffusion (CLD-SGM) \cite{dockhorn2021score} augments the data dimensions with momentum dimensions and diffuses only in momentum space, resulting in more Gaussian scores and fewer evaluations for comparable quality. However, both these methods significantly modify the original formulation of score-based models, such that exact likelihood evaluation and controllable generation become considerably more difficult.\footnote{In CLD-SGM, one must marginalize over the momentum variables; and in LSGM one must marginalize over the latent variable of VAE.}

Unlike these previous works, subspace diffusion simultaneously improves sample quality and inference runtime while also preserving all the capabilities of the original formulation. Compared with LSGM and CLD-SGM, subspace diffusion also has the advantage of being compatible with existing trained score models, incurring only the overhead required to train the smaller subspace score models.

\paragraph{\textbf{Cascading generative models.}} Subspace diffusion bears some similarity to cascading generative models consisting of one low-dimensional model followed by one or more super-resolution models \cite{menick2018generating,razavi2019generating}. Cascading score-based models have yielded strong results on high-resolution class-conditional ImageNet generation \cite{dhariwal2021diffusion,saharia2021image,ho2021cascaded}. These models formulate each super-resolution step as a full diffusion process conditioned on the lower-resolution image. Subspace diffusion, on the other hand, models a single diffusion process punctuated by projection steps. This leads to a more general theoretical framework that is useful even in domains where the concept of super-resolution does not apply (see for example the synthetic experiments in Appendix~\ref{app:synthetic}). Chaining conditional diffusion processes also complicates the application of other capabilities of score-based models---for example, evaluating log-likelihoods would require marginalizing over the intermediate lower-resolution images. Our subspace diffusion framework is a modification of a single diffusion and does not incur these difficulties.

\section{Subspace Diffusion} \label{sec:subspace}
A concrete formulation of a score-based model requires a choice of forward diffusion process, specified by $\mathbf{f}(\mathbf{x}, t)$, $\mathbf{G}(\mathbf{x}, t)$. Almost always, these are chosen to be \emph{isotropic}, i.e., of the form 
\begin{equation} \label{eq:isotropic}
    \mathbf{f}(\mathbf{x}, t) = f(t)\, \mathbf{x} \quad \mathbf{G}(\mathbf{x}, t) = g(t) \, \mathbf{I}_d
\end{equation} where $d$ is the data dimensionality. For example, the variance exploding (VE) SDE has $f(t)=0$ and $g(t) = \sqrt{d\sigma^2/dt}$ where $\sigma^2(t)$ is the variance of the perturbation kernel at time $t$ \cite{song2021score}. The sole exception is the Langevin diffusion in CLD-SGM \cite{dockhorn2021score}, but this required new forms of score-matching and specialised SDE solvers for numerical stability. We aim to keep the simplicity and convenience of form \eqref{eq:isotropic} while addressing its limitations discussed in Section~\ref{sec:intro}. We thus propose that at every point in time, the diffusion is restricted \emph{to some subspace}, but is otherwise isotropic \emph{in that subspace}. Specifically, the forward diffusion begins in the full space, but is projected and restricted to increasingly smaller subspaces as time goes on. Any isotropic forward diffusion can therefore be converted into a subspace diffusion.

For any diffusion with the form \eqref{eq:isotropic}, define the corresponding subspace diffusion as follows. Divide $(0, T)$ into $K+1$ subintervals, $(t_0, t_1), \ldots, (t_K, t_{K+1})$ where for notational convenience $t_0=0, t_{K+1}=T$. Then define:
\begin{equation} \label{eq:G}
    \mathbf{G}(\mathbf{x}, t) = g(t)\mathbf{U}_k\mathbf{U}_k^T    
\end{equation}
for each interval $t_k < t < t_{k+1}$, where $\mathbf{U}_k \in \mathbb{R}^{d \times n_k}$ is the matrix whose $n_k \le d$ orthonormal columns span a subspace of $\mathbb{R}^d$. We refer to this subspace as the $k$th subspace and to the columns of $\mathbf{U}_k$ as its basis. For notational convenience, $\mathbf{U}_0 = \mathbf{I}_d$. We choose $n_k$ such that $d = n_0 > n_1 > \ldots > n_K$. We also require the $k$th subspace to be a subspace of the $j$th subspace for any $j < k$, which can be written as $\mathbf{U}_{j}\mathbf{U}_{j}^T\mathbf{U}_{k}=\mathbf{U}_{k}$. Together, these definitions state that diffusion is coupled or constrained to occur in progressively smaller subspaces defined by $\mathbf{U}_k$ in the interval $(t_k, t_{k+1})$.

Turning to $\mathbf{f}(\mathbf{x}, t)$, define 
\begin{equation} \label{eq:f}
    \mathbf{f}(\mathbf{x}, t) = f(t)\,\mathbf{x} + \sum_{k=1}^K \delta(t-t_k)(\mathbf{U}_k\mathbf{U}_k^T - \mathbf{I}_d)\,\mathbf{x}
\end{equation}
where $\delta$ is the Dirac delta. This states that at time $t_k$, $\mathbf{x}$ is projected onto the $k$th subspace. Figure~\ref{fig:method} illustrates the high-level idea of subspace diffusion, along with some of its properties discussed in more detail below.

\paragraph{\textbf{Notation.}} For the rest of the exposition, we define:
\begin{itemize}
    \item $\mathbf{U}_{k \mid j} = \mathbf{U}_j^T\mathbf{U}_k \in \mathbb{R}^{n_j \times n_k}$ for $j \le k$ defines the $k$th subspace written in the basis of the $j$th subspace. In particular, $\mathbf{U}_{k \mid 0} = \mathbf{U}_k$ and $\mathbf{U}_{k \mid k} = \mathbf{I}_{n_k}$.
    \item $\mathbf{P}_{k \mid j} = \mathbf{U}_{k\mid j}\mathbf{U}_{k \mid j}^T \in \mathbb{R}^{n_j \times n_j}$ for $j \le k$ is the projection operator onto the $k$th subspace, written in the basis of the $j$th subspace.
    \item $\mathbf{P}_{k \mid j}^\perp = \mathbf{I}_{n_j} - \mathbf{P}_{k \mid j}  \in \mathbb{R}^{n_j \times n_j}$ for $j < k$ is the projection operator onto the complement of the $k$th subspace, written in the basis of the $j$th subspace.
    \item $\mathbf{x}_{k} = \mathbf{U}_k^T\mathbf{x} \in \mathbb{R}^{n_k}$ is the component of $\mathbf{x}$ in the $k$th subspace, written in that basis. In particular, $\mathbf{x}_0 = \mathbf{x}$.
    \item $\mathbf{x}^\perp_{k\mid j} = \mathbf{P}^\perp_{k\mid j}\mathbf{x}_j \in \mathbb{R}^{n_j}$ for $j < k$ is the component of $\mathbf{x}_j$ orthogonal to the $k$th subspace, written in the basis of the $j$th subspace.
\end{itemize}

\subsection{Score matching} \label{sec:score}
To generate data, we need to learn the score $\nabla_\mathbf{x} \log p(\mathbf{x}, t)$ as usual. However, for times $t_k < t < t_{k+1}$, the support of $p(\mathbf{x}, t)$ is only in the $k$th subspace. This means that if we learn a separate score model $\mathbf{s}_k(\mathbf{x}, t) \approx \nabla_\mathbf{x} \log p(\mathbf{x}, t)$ for each interval $t \in (t_k, t_{k+1})$, then the model $\mathbf{s}_k$ \emph{only needs to have dimensionality} $n_k$. In particular, we use models smaller than $n_0 = d$ for all times $t > t_1$.

To learn these lower-dimensional models, we leverage the fact that the subspace components $\mathbf{x}_k$ of the data diffuse under an SDE with the same $f(t), g(t)$ as the full data, independent of the orthogonal components. This is due to the fact that the original diffusion is isotropic. To see this, consider (for simplicity) the case $K=1$, i.e., we only use one proper subspace. Then since $ d\mathbf{x}_1 = \mathbf{U}_1^Td\mathbf{x}$,

\begin{equation}
\begin{aligned}
    d\mathbf{x}_1 &= f(t)\mathbf{U}_1^T\mathbf{x} \; dt + \delta(t-t_1)\mathbf{U}_1^T(\mathbf{U}_1\mathbf{U}_1^T - \mathbf{I}_d)\mathbf{x} \; dt \\
    &\quad + g(t)\left(\mathbf{U}_1^T\left(\mathbbm{1}_{t < t_1}\mathbf{I}_d + \mathbbm{1}_{t > t_1}\mathbf{U}_1\mathbf{U}_1^T\right)\right) \; d\mathbf{w}
\end{aligned}
\end{equation}
However, because $\mathbf{U}_1^T\mathbf{U}_1 = \mathbf{I}_d$, the above simplifies as
\begin{equation}
\begin{aligned}
    d\mathbf{x}_1 
    &= f(t)\mathbf{x}_1 \; dt + g(t) \; d\mathbf{w}_1
\end{aligned}
\end{equation}
where, because the columns of $\mathbf{U}_1$ are orthonormal, $d\mathbf{w}_1 := \mathbf{U}_1^T \; d\mathbf{w}$ is a Brownian diffusion in $\mathbb{R}^{n_1}$. As a result, the perturbation kernels $p(\mathbf{x}_1(t) \mid \mathbf{x}_1(0))$ in the subspace have the same form as in the full space. This allows us to train a model to match the scores $\nabla_{\mathbf{x}_1} \log p(\mathbf{x}_1, t)$ via precisely the same procedure as in \cite{song2021score}, except we treat $\mathbf{x}_1(0)$ as the original undiffused data. These scores are related to the full-dimensional scores $\nabla_\mathbf{x} \log p(\mathbf{x}, t)$ via $\mathbf{U}_1$, but since $\mathbf{x} = \mathbf{U}_1\mathbf{x}_1$ for times $t > t_1$, we can directly work with data points $\mathbf{x}_1$ and score models $\nabla_{\mathbf{x}_1} \log p(\mathbf{x}_1, t)$ with no loss of information for times $t>t_1$. Thus, in the general case, we train $K+1$ different score models $\mathbf{s}_k(\mathbf{x}_k, t) \approx \nabla_{\mathbf{x}_k} \log p(\mathbf{x}_k, t)$, where we consider $\mathbf{x}_k$ to have diffused under the original $f(t), g(t)$ for the full time scale $(0, T)$.

\subsection{Sampling} \label{sec:sampling}

To generate a sample, we use each score model $\mathbf{s}_k(\mathbf{x}_k, t)$ in the corresponding interval $(t_k, t_{k+1})$ to solve the reverse diffusion of $\mathbf{x}_k$. However, we cannot use the score to reverse the projection steps at the boundaries times $t_k$. Thus, to impute $\mathbf{x}_{k-1}(t_{k})$ from $\mathbf{x}_{k}(t_{k})$, we sample $\mathbf{x}^\perp_{k \mid k-1}(t_k)$ by injecting isotropic Gaussian noise orthogonal to the $k$th subspace. The variance $\Sigma^\perp_{k\mid k-1}$ of the injected noise is chosen to match the marginal variance of $\mathbf{x}^\perp_{k \mid k-1}$ at time $t_k$, which is the sum of the original variance of $\mathbf{x}^\perp_{k \mid k-1}$ in the data and the variance of the perturbation kernel:
\begin{equation} \label{eq:injection}
\begin{aligned}
    \Sigma^\perp_{k \mid k-1}(t_{k}) := \frac{\alpha^2(t_k)}{n_{k-1} - n_{k}} \mathbb{E}\left[ \lVert \mathbf{x}^\perp_{k\mid k-1}(0) \rVert^2_2 \right] + \sigma^2(t_k)
\end{aligned}
\end{equation}
where $\alpha(t)$ and $\sigma^2(t)$ are the scale and variance of the perturbation kernels.

Sampling $\mathbf{x}^\perp_{k\mid k-1}$ in this manner assumes that (at time $t_k$) it is independent of $\mathbf{x}_{k}$ and roughly an isotropic Gaussian. The final sample quality will depend on the validity of this assumption. Intuitively, however, we specifically choose subspaces and times such that the original magnitude of $\mathbf{x}_{k\mid k-1}^\perp$ (the first term in \eqref{eq:injection}) is very small compared to the diffusion noise (the second term), which is indeed isotropic and independent of the data. We also find that a few conditional Langevin dynamics steps with $s_{k}(\mathbf{x}_k, t_{k+1})$ to correct for the approximations of noise injection help sampling quality. The complete sampling procedure is outlined in Algorithm~\ref{alg:sampling}.

So far we have presented subspace diffusion as an explicit modification to the forward diffusion involving projection and confined diffusion, which best matches how we implement unconditional sample generation. However, an alternate view is more suitable for controlled generation, where a full-dimensional score model is required; or in ODE-based likelihood evaluation or probability flow sampling, where the adaptive, non-monotonic evaluations make working with discrete projection steps inconvenient. In these settings, we regard subspace diffusion at time $t \in (t_k, t_{k+1})$ as \emph{explicitly} modeling the score component in $k$th subspace with $\mathbf{s}_k(\mathbf{x}_k, t)$, and \emph{implicitly} modeling all orthogonal components with Gaussians. Specifically, for $t \in (t_k, t_{k+1})$ we decompose $\mathbf{x}$ as
\begin{equation} \label{eq:decomp}
    \mathbf{x} = \sum_{j=0}^{k-1} \mathbf{U}_j \mathbf{x}^\perp_{j+1 \mid j} + \mathbf{U}_k\mathbf{x}_k
\end{equation}
where the sum corresponds to the components that are "Gaussianized" out by each projection step. We thus model each $\mathbf{x}^\perp_{j+1 \mid j}$ implicitly as isotropic Gaussian with variance $\Sigma^\perp_{j+1\mid j}$, and model $\mathbf{x}_k$ explicitly with score model $\mathbf{s}_k(\mathbf{x}_k, t)$, giving the full score:
\begin{equation} \label{eq:full-score}
    \nabla_\mathbf{x}\log p(\mathbf{x}, t) \approx \mathbf{U}_k\mathbf{s}_k(\mathbf{U}_k^T\mathbf{x}, t) - \sum_{j=0}^{k-1} \left(\mathbf{P}_{j\mid 0}-\mathbf{P}_{j+1 \mid 0}\right)\frac{\mathbf{x}}{\Sigma^\perp_{j+1\mid j}(t)}
\end{equation}
where, for clarity, we write all components in terms of $\mathbf{x}$ in the original basis.

\begin{algorithm}[t]
\caption{Unconditional sampling with subspace diffusion}\label{alg:sampling}
\KwIn{subspaces $\mathbf{U}_k$, projection times $t_k$, score models $\mathbf{s}_k(\mathbf{x}_k, t)$, $k = 0 \ldots K$}
\KwOut{approximate sample $\mathbf{x}_0$ from $p(\mathbf{x}_0, 0) = p_\text{data}(\mathbf{x})$}
$\mathbf{x}_K \gets $ sample from prior $p(\mathbf{x}_K, T) \in \mathbb{R}^{n_K}$\;
\For{$k\leftarrow K$ \KwTo $0$}{
  $\mathbf{x}_k \gets $ solve reverse SDE with $\mathbf{s}_k(\mathbf{x}_k, t)$ from $t_{k+1}$ to $t_k$ starting from $\mathbf{x}_k$\;
  \If{$k > 0$}{
    $\mathbf{x}^\perp_{k\mid k-1} \gets $ sample from $\mathcal{N}(\mathbf{0}, \Sigma^\perp_{k\mid k-1}(t_k) \; \mathbf{I}) \in \mathbb{R}^{n_{k-1}}$  \;
    $\mathbf{x}^\perp_{k\mid k-1} \gets \mathbf{P}^\perp_{k\mid k-1}\mathbf{x}^\perp_{k\mid k-1} $ \;
    $\mathbf{x}_{k-1} \gets \mathbf{U}_{k\mid k-1}\mathbf{x}_k + \mathbf{x}^\perp_{k\mid k-1}$ \;
    \For(\tcp*[f]{n is a hyperpameter}){$i\leftarrow 1$ \KwTo $n$}{
        $\mathbf{x}_{k-1} \gets \text{LangevinStep}(\mathbf{x}_{k-1}, t_k)$  \;
    }
  }
}
\end{algorithm}

\subsection{Image subspaces}

We now restrict our attention to generative modeling of natural images.\footnote{See Appendix~\ref{app:synthetic} for experiments on more generic synthetic data.} Motivated by the observation that adjacent pixels tend to be \emph{similar} in color, we choose subspaces that correspond to images where adjacent groups of pixels are \emph{equal} in color---i.e., downsampled versions of the image. Henceforth, we refer to such \emph{downsampling subspaces} in terms of their resolution (e.g., the $16\times 16$ subspace), refer to projection onto subspaces at times $t_k$ as \emph{downsampling}, and to the reverse action as \emph{upsampling}.\footnote{It is via this choice of subspace that subspace diffusion superficially resembles the cascading models discussed in Section~\ref{sec:background}.}

To more precisely formulate these subspaces, suppose we have a full-resolution image $\mathbf{X} \in \mathbb{R}^{(n \times n \times 3)}$. In particular, we will work with $n$ that are integer powers of 2. Then we define a downsampling operator $\mathcal{D}: \mathbb{R}^{(n \times n \times 3)} \rightarrow \mathbb{R}^{(n/2 \times n/2 \times 3)}$ such that if $\mathbf{X}_{k+1} = \mathcal{D}\mathbf{X}_k$, then
\begin{equation} \label{eq:downsample}
    \mathbf{X}_{k+1}[a, b, c] = \frac{1}{2}\sum_{(i,j)\in \{0,1\}^2} \mathbf{X}_k[2a+i, 2b+j, c]
\end{equation}
which states that $\mathbf{X}_{k+1}^{(t)}$ is simply $\mathbf{X}_k^{(t)}$ after mean-pooling $2 \times 2$ patches, multiplied by $2$. We can use $\mathcal{D}$ to implicitly define $\mathbf{U}_k$:
\begin{equation} \label{eq:uk}
    \mathbf{U}_k^T \mathbf{x} = \mathcal{D}^k\mathbf{x} \quad \text{or} \quad \mathbf{U}_k^T \mathbf{x} = \mathcal{D}\mathbf{U}_{k-1}^T\mathbf{x}
\end{equation}
where here we consider $\mathbf{x}$ to be the column vector representation of the array $\mathbf{X}$. The choice of $\mathcal{D}$ corresponds to orthonormal $\mathbf{U}_k$, as each column of $\mathbf{U}_k$ has $2^{2k}$ nonzero entries, each with magnitude $1/2^k$. Thus, all of the general results from the preceding section apply. In particular, we can consider the same forward diffusion process defined by $f(t), g(t)$ to be occurring for each downsampled image $\mathcal{D}^k\mathbf{x}$, such that the subspace score models  $\mathbf{s}_k(\mathbf{x}_k, t)$ correspond to the same score model trained over a {lower-resolution} version of the same dataset.

It is natural to consider whether there may exist more optimal subspaces for natural images. In Table~\ref{tab:pca} we compare the downsampling subspaces to the optimal subspaces of equivalent dimensionality\footnote{That is, an $N \times N$ subspace has dimensionality $3N^2$.} found by principle components analysis (PCA) in terms of root mean square distance (RMSD) of the data from the subspace. Generally, the downsampling subspaces can be seen to be suboptimal. However, if we were to use the optimal PCA subspaces, the coordinates would not take the form of an image-structured latent with translation equivariance, and thus would be incompatible with the convolutional neural networks in the score model. Therefore, a more appropriate comparison is with the subspace found via PCA of the distribution of all \emph{patches} of pixels of the appropriate size, which we call Patch-PCA (see Appendix~\ref{appendix:patch-pca} for details). These subspaces offer only minor improvements over the downsampling subspaces, so we did not explore them further.

It is also possible that for any given dimensionality $n < d$, the $n$-dimensional substructure that best approximates the data distribution is a nonlinear manifold rather than a subspace. However, leveraging such manifolds to reduce the dimensionality of diffusion would require substantial modifications to the present framework. While potentially promising, we leave such extensions to nonlinear manifolds to future work.

\begin{table}[t!]
    \centering
    \begin{tabular}{ccccc}  
    \toprule
    & & \multicolumn{3}{c}{RMSD per dim.}\\ \cmidrule{3-5}
    Dataset  & Subspace dim. & PCA & Patch-PCA & Downsampling \\ \midrule
    \multirowcell{2}{CIFAR-10 \\ ($32\times 32$)}   
    & $16\times 16$ & 0.024 & 0.064 & 0.075\\
    & $8 \times 8$  & 0.061 & 0.093 & 0.110\\    \midrule
    \multirowcell{3}{CelebA-HQ \\ ($256\times 256$)}
    & $128\times 128$ & --- & 0.034 & 0.034\\
    & $32 \times 32$  & 0.041 & 0.063 & 0.073\\
    & $8 \times 8$  & 0.083 & 0.117 & 0.141\\ 
    \midrule
    \multirowcell{3}{LSUN Church \\ ($256\times 256$)}
    & $128\times 128$ & --- & 0.058 & 0.070\\
    & $32 \times 32$  & 0.082 & 0.099 & 0.109\\ 
    & $8 \times 8$  & 0.126 & 0.146 & 0.158\\ 
    \bottomrule
    \end{tabular}
    \caption{Comparison of downsampling subspaces with optimal subspaces of equivalent dimensionality found by PCA and Patch-PCA. RMSD per dim refers to RMSD$/\sqrt{d-n}$, where $d, n$ are the original and subspace dimensionalities. PCA and Patch-PCA were run on a subset of CelebA and LSUN.}
    \label{tab:pca}
\end{table}

\subsection{Orthogonal Fisher divergence} \label{sec:fisher}

We now propose a principled manner to choose among the candidate subspaces for a given image dataset, as well as the downsampling times $t_k$.

For any fixed choice of proper subspaces $\mathbf{U}_1\ldots\mathbf{U}_k$, the optimal values of each $t_k$ must balance two factors: smaller $t_k$ reduces the number of reverse diffusion steps occurring at higher dimensionality $n_{k-1}$, whereas larger $t_k$ makes the Gaussian approximation of the orthogonal components $\mathbf{x}^{\perp}_{k\mid k-1}$ more accurate when we sample at time $t_k$. This suggests that we should choose the minimum times that keep the error of the Gaussian approximation below some tolerance threshold. However, we cannot quantify the true error as we do not have access to the underlying distribution of $\mathbf{x}^\perp_{k\mid k-1}$. Thus, we instead examine how much the \emph{learned} full-dimensional score model $\mathbf{s}_{0}(\mathbf{x}, t)$ diverges from the Gaussian approximation on $\mathbf{x}^\perp_{k\mid k-1}$ as $t$ is varied. To quantify this divergence, for any $j < k$ we introduce the \emph{orthogonal Fisher divergence} of $\mathbf{U}_{k\mid j}$ as:
\begin{equation}\label{eq:fisher}
    D_F(\mathbf{U}_{k\mid j}; t) = \frac{\Sigma^\perp_{k\mid j}(t)}{n_{j} - n_{k}}\mathbb{E}_{\mathbf{x}(t)}\left[\left\lVert \mathbf{P}^\perp_{k\mid j}\mathbf{U}_{j}^T\mathbf{s}_{0}(\mathbf{x}, t) + \frac{\mathbf{x}_{k\mid j}^\perp}{\Sigma^\perp_{k\mid j}(t)} \right\rVert^2 \right]
\end{equation}
The first term is the component of the score orthogonal to $\mathbf{U}_{k\mid j}$, and the second term is the score of the Gaussian approximation of $\mathbf{x}^\perp_{k\mid j}$. The divergence is normalized by the (approximate) expected norm of the Gaussian score, which enables values for different $t, j, k$ to be compared. The expectation over $\mathbf{x}(t)$ can then be approximated using the training data. The divergence $D(\mathbf{U}_{k\mid k-1}; t)$ then corresponds to the error that would be introduced by the upsampling step at time $t$.

\begin{figure}[t!]
    \centering
    \includegraphics[width=\textwidth]{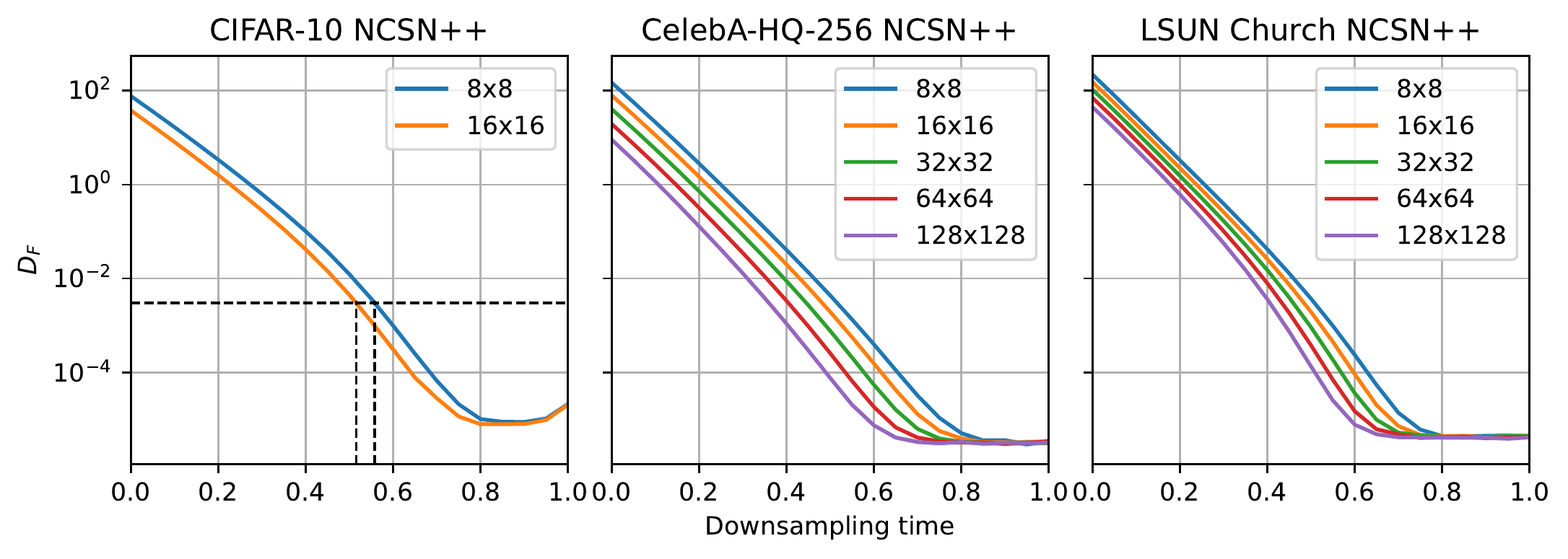}
    \caption{Orthogonal Fisher divergence plots computed with respect to the pretrained NCSN++ full-dimensional score models from \cite{song2021score}. Similar plots can be generated for other models. Given a divergence threshold, the optimal downsampling times $t_k$ for any subspace sequence are the times at which the corresponding divergences attain that threshold. For example, on CIFAR-10 with a target $D_F = 3\times 10^{-3}$ and the sequence $32 \rightarrow 16 \rightarrow 8$, the downsampling times are $t_1=0.516, t_2=0.558$. In this case, the intermediate $16\times 16$ subspace would be used for only $4.2\%$ of the diffusion. As the plot shows, this imbalance would characterise any sequence of more than one proper subspace.}
    \label{fig:fisher}
\end{figure}

Given a sequence of subspaces, the divergence threshold becomes the sole hyperparameter of the sampling process, as we can compute \eqref{eq:fisher} to determine the upsampling times $t_k$ for any threshold. Once the $t_k$ are known, we can estimate the runtime improvement over the full-dimensional score model. Thus, we can choose the subspaces sequence to minimize the estimated runtime. Additionally, it is more convenient to consider $D_F(\mathbf{U}_{k\mid 0}; t)$ as opposed to $D(\mathbf{U}_{k\mid k-1}; t)$, which corresponds to assuming that at time $t_k$, $\mathbf{x}_{k\mid k-1}^\perp$ is sampled with variance $\Sigma^\perp_{k\mid 0}$ rather than $\Sigma^\perp_{k\mid k-1}$.\footnote{The difference is minimal as the variance of the perturbation kernel dominates either term for reasonable divergence thresholds.} The benefit of this approximation is that we can speak of the divergence purely as a property of the subspace, independent of the preceding subspace (if any). Thus, we can simultaneously plot the orthogonal Fisher divergence for each downsampling subspace, as illustrated in Figure~\ref{fig:fisher}. The choice of intervals for any subspace sequence and divergence threshold can then be directly read off the plot.

As Figure~\ref{fig:fisher} shows, for standard image datasets there appears to be little utility to using more than one proper subspace, as the diffusion in intermediate dimensions would be very brief. On the other hand, training additional models is computationally expensive and adds to the sum of the model sizes required for inference. Thus, our experiments focus on subspace diffusions consisting of only one proper downsampling subspace. In particular, for CIFAR-10, we consider the $8 \times 8$ and $16 \times 16$ subspaces separately, while for CelebA-HQ and LSUN Church we consider only the $64 \times 64$ subspace, which offers the best potential runtime improvement.

\section{Experiments} \label{sec:experiments}

We demonstrate the utility and versatility of our method by improving upon and accelerating state-of-the-art continuous score-based models. Specifically, we take the pretrained models on CIFAR-10, CelebA-256-HQ, and LSUN Church from \cite{song2021score} as full-dimensional score models, train additional subspace score models of the same architecture, and use them together in the subspace diffusion framework. All lower-dimensional models are trained with the same hyperparameters and training procedure as the original model (see Appendix~\ref{appendix:hyperparameters}). During inference, we use the unmodified reverse SDE solvers and the same number and spacing of denoising steps. We investigate results for a range of divergence thresholds, corresponding to different durations of diffusion in the subspace.

For all experiments, further results and additional samples may be found in Appendix~\ref{appendix:detailed-results} and Appendix~\ref{appendix:samples}, respectively.

\begin{table}[t!] 
\begin{minipage}[b]{.32\linewidth}
    \centering
    {\small
    \begin{tabularx}{\textwidth}{lY}
    \toprule
    Model                           & FID $\downarrow$   \\ \midrule
    DDIM  \cite{song2020denoising}  & 4.04 \\ 
    FastDPM \cite{kong2021fast} & 2.86  \\ 
    Bilateral DPM \cite{lam2021bilateral} & 2.38  \\ %\midrule
    Analytic DPM \cite{bao2022analytic} & 3.04  \\ 
    Prog. Distillation \cite{salimans2021progressive} & 2.57  \\
    CLD-SGM \cite{dockhorn2021score} & 2.23  \\
    LSGM \cite{vahdat2021score} & 2.10 \\
    Adaptive solver \cite{jolicoeur2021gotta} & 2.44  \\
     \bottomrule
    \end{tabularx}   
    }
\end{minipage} \hspace{0.1cm}
\begin{minipage}[b]{.67\linewidth}
    \centering
    {\small
    \begin{tabularx}{\textwidth}{llYYYYY}
    \toprule
    Model  && FID $\downarrow$   & IS $\uparrow$ & Thresh.   & $t_1$ & Run. \\ \midrule
    
    \multirow{2}{*}{NCSN++}     
    & full       & 2.38 & 9.83 \\ 
    & subspace   & 2.29 & \textbf{9.99}	& 3e-3 & 0.52 & 0.66 \\ 
    \multirow{2}{*}{\shortstack[l]{NSCN++\\(deep)}}
    & full       & 2.20 & 9.89 \\ 
    & subspace  & \textbf{2.17} & 9.94	& 1e-3 & 0.56 & 0.69 \\ 
    \multirow{2}{*}{DDPM++}
    & full       & 2.61 & 9.56 \\ 
    & subspace   & 2.60 & 9.54 & 3e-5 & 0.62 & 0.73 \\ 
    \multirow{2}{*}{\shortstack[l]{DDPM++\\(deep)}}
    & full       & 2.41  & 9.57 \\ 
    & subspace   & 2.40 & 9.66 & 1e-4 & 0.56 & 0.69 \\ \bottomrule
    \end{tabularx}   
    }
\end{minipage}
\caption{CIFAR-10 sample quality for 50k images.  \textit{Left}: the best performance of previous methods to accelerate score-based models. \textit{Right}: the original full diffusion from \cite{song2021score} and the respective best subspace diffusion (all $16\times 16$), with the corresponding divergence threshold, downsampling time $t_1$, and empirical runtime relative to the full model.}
    \label{tab:results}
    \vspace{-10pt}
\end{table}
\begin{figure}[t]
    \centering
    \includegraphics[width=0.9\textwidth]{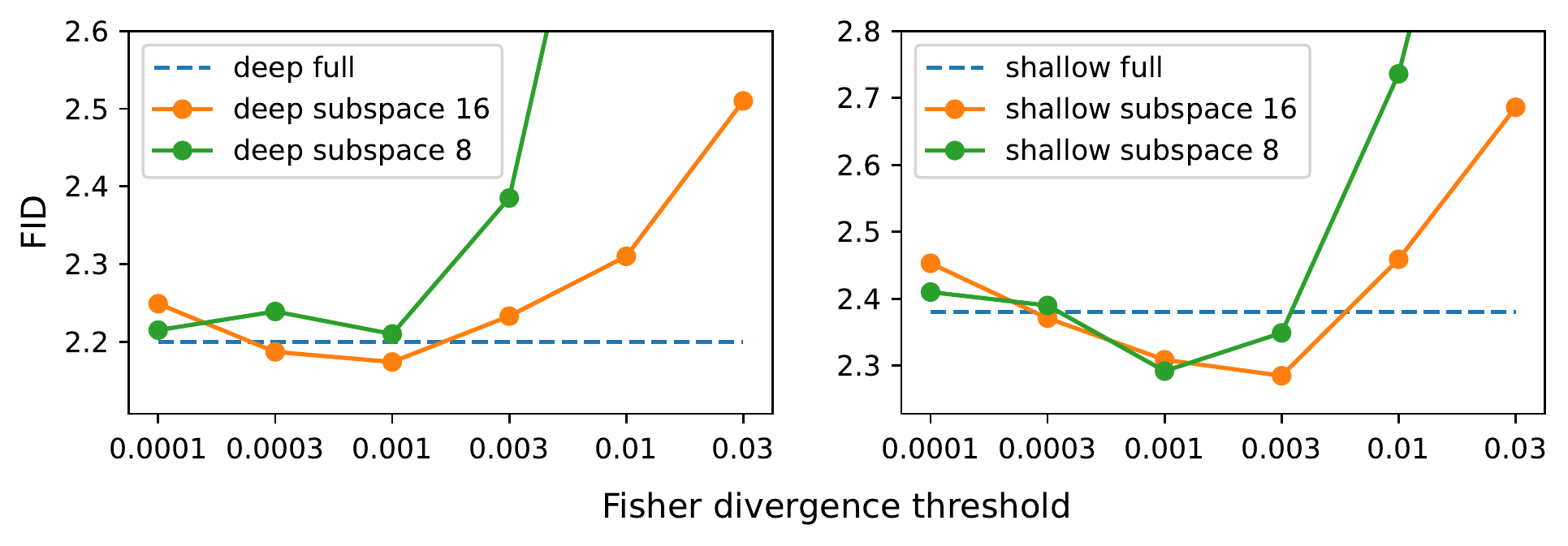}
    \caption{CIFAR-10 sample quality from NCSN++ subspace diffusion (shallow and deep models) with different subspaces and divergence thresholds.}
    \label{fig:threshold}
        \vspace{-10pt}
\end{figure}

\paragraph{\textbf{Unconditional sampling}} We evaluate subspace diffusion on unconditional CIFAR-10 generation with the Inception score (IS) and distance (FID) as metrics. We examine both the NCSN++ and DDPM++ models from \cite{song2021score}, which correspond to different forward diffusion processes, as well as the deep versions of these models, for a total of 4 full-dimensional models. For each model, we separately construct subspace diffusion with $8\times 8$ and $16\times 16$ subspaces. As in \cite{song2021score}, we choose the best checkpoint by FID.

In Figure \ref{fig:threshold}, we show the performance of the NCSN++ subspace diffusion models for different choices of the Fisher divergence threshold $D_F$. In all cases, the models display U-shaped performance curves as the threshold is varied. When the threshold is small, most of the diffusion is done at full dimensionality, and the performance is close to that of the full model alone. As the threshold increases and more diffusion is done in the subspace, the models \emph{improve} over the full model until reaching the best performances between $D_F =1 \times 10^{-3}$ and $D_F = 3\times 10^{-3}$. This improvement offers support for the hypothesis, discussed in the introduction, that restricting the dimensionality (or support) of the score to be matched can help the subspace model learn and extrapolate more accurately than the full-dimensional model. Finally, for large thresholds the performance deteriorates as the Gaussian approximation of the orthogonal component becomes too inaccurate.

Table \ref{tab:results} compares the performance of the best subspace diffusion models with the original full-dimensional models from \cite{song2021score} and with prior methods for accelerating score-based models. Subspace diffusion and LSGM \cite{vahdat2021score} are the only methods where the improved runtime does not come at the cost of decreased performance (relative to \cite{song2021score}). The runtime improvement over the full-dimensional baseline varies with the choice of divergence threshold; for those leading to the best sample qualities, the improvements are typically around 30\%. Since the concept of subspace diffusion is orthogonal to the techniques used by most previous work (see Section \ref{sec:background}), it can potentially be used in combination with them for further runtime improvement. 

\begin{figure}[t]
    \centering
    \includegraphics[width=\textwidth]{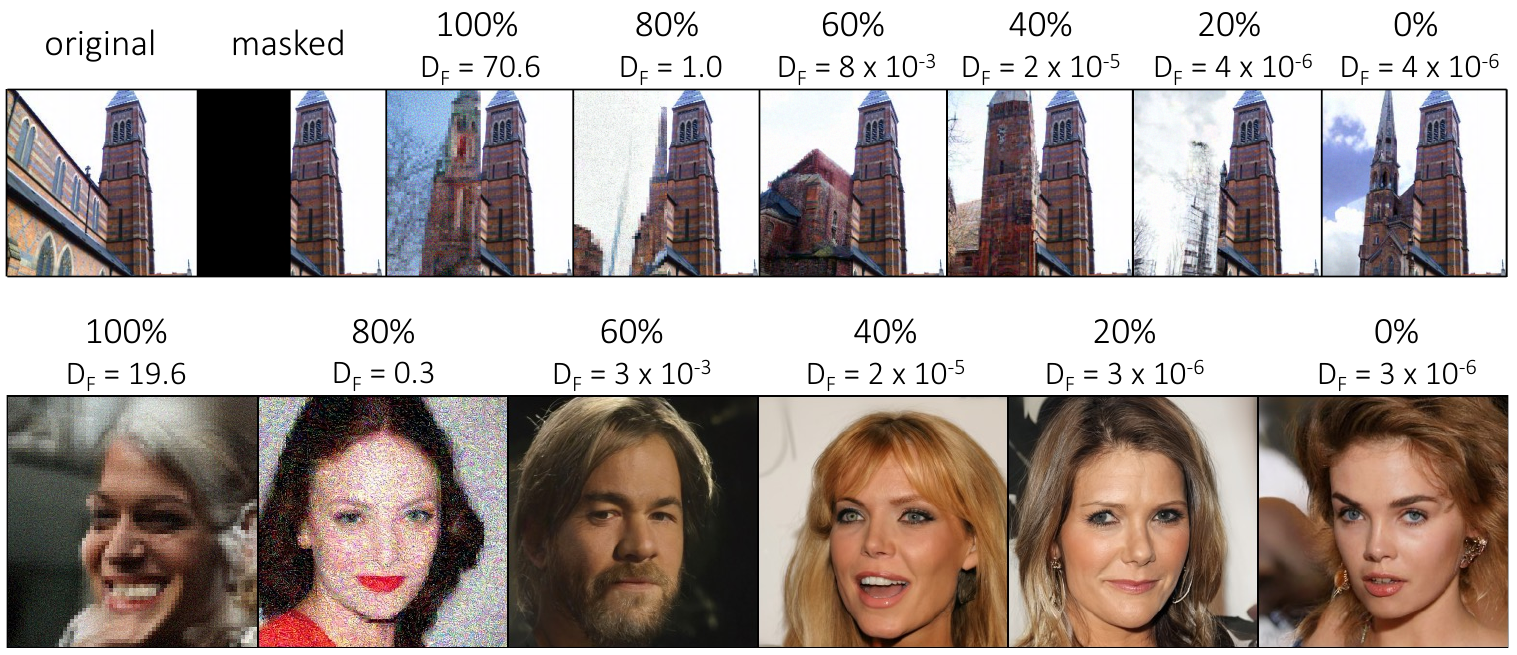}
    \caption{Random high resolution samples with $64\times 64$ subspace diffusion. \textit{Top:} Inpainting on the $256 \times 256$ LSUN Church dataset. \textit{Bottom:} Unconditional generation of samples for CelebA-HQ-256. From right to left, the fraction of the diffusion in the subspace increases in intervals of 20\%, with the corresponding orthogonal Fisher divergence shown. As expected from the divergence analysis in Figure~\ref{fig:fisher}, there is little deterioration in quality for images generated with up to 60\% of the diffusion in the subspace.} 
    \label{fig:high_res}
\end{figure}

Next, we show the applicability of our method to higher-resolution datasets by generating samples on CelebA-HQ-256 with NCSN++ subspace diffusion. As discussed in Section \ref{sec:fisher}, we use only the $64\times 64$ subspaces and perform no hyperparameter tuning or checkpoint selection. In Figure~\ref{fig:high_res}, we show random samples from CelebA-HQ for different amounts of diffusion in the subspace, along with the corresponding Fisher divergence. Qualitatively, we can restrict up to 60--70\% of the diffusion to the subspace without significant loss of quality.

\begin{wraptable}[13]{r}{0.45\textwidth}
    \vspace{-12pt}
    \small%    \footnotesize
    \begin{tabular}{lccc}
    \toprule
    Subspace               & Thresh. & NLL $\downarrow$   & FID $\downarrow$   \\ \midrule
    None & --- & 2.995 & 2.95 \\ \midrule
    \multirow{3}{*}{$8\times 8$} 
    & $1 \times 10^{-4}$ & 2.998 & 3.02 \\
    & $3 \times 10^{-4}$ & 2.999 & 3.12 \\
    & $1 \times 10^{-3}$ & 2.998 & 3.53 \\ \midrule
    \multirow{3}{*}{$16\times 16$} 
    & $1 \times 10^{-4}$ & 2.997 & 2.95 \\
    & $3 \times 10^{-4}$ & 2.997 & 3.00 \\
    & $1 \times 10^{-3}$ & 2.997 & 3.17 \\ \bottomrule
    \end{tabular}
    \caption{ODE sampling and NLL evaluation on CIFAR-10 from DDPM++ (deep) with subspace diffusion.}
    \label{tab:ode_results}
\end{wraptable}
\paragraph{\textbf{ODE sampling and likelihood}} Subspace diffusion retains the flexible capabilities of the continuous-time SDE framework. In particular, the corresponding probability flow ODE \eqref{eq:ode} can be used to evaluate exact log-likelihoods and generate samples, as described in \cite{song2021score}. In Table \ref{tab:ode_results}, we show results for these tasks on CIFAR-10 for subspace diffusion in combination with the DDPM++ (deep) model. We use the alternate subspace score formulation \eqref{eq:full-score} with the original ODE solvers, and use the last checkpoint of each training run. Subspace diffusion has little to no impact on the log-likelihoods obtained and slightly hurts sample quality.

\paragraph{\textbf{Inpainting}} Subspace diffusion can also be used for controllable generation tasks, an example of which is image inpainting. Indeed, by using the alternate formulation \eqref{eq:full-score}, the subspace model appears as a full-dimensional model and integrates seamlessly with the existing inpainting procedures described in \cite{song2021score}. In Figure \ref{fig:high_res}, we show inpainting results on LSUN Church with $64\times 64$ subspace diffusion in conjunction with the pretrained NCSN++ model. As with the unconditional samples, quality does not significantly deteriorate with up to 60\% of the diffusion occurring in the subspace.

\section{Conclusion}
We presented a novel method for more efficient generative modeling with score-based models. \textit{Subspace diffusion models} restrict part of the diffusion to lower-dimensional subspaces such that the score of the projected distribution is faster to compute and easier to learn. Empirically on image datasets, our method provides inference speed-ups while preserving or improving the performance and capabilities of  state-of-the-art models. Potential avenues of future work include applying subspace diffusion to other data domains and combining it with step-size based methods for accelerating inference. More generally, we hope that our work opens up further research on dimensionality reduction in diffusion processes, particularly to nonlinear manifolds and/or learned substructures, as a means of both simplifying and improving score-based generative models.

\section*{Acknowledgments}
We thank Yilun Du, Xiang Fu, Jason Yim, Shangyuan Tong, Yilun Xu, Felix Faltings, and Saro Passaro for helpful feedback and discussions. Bowen Jing acknowledges support from the Department of Energy Computational Science Graduate Fellowship.

\bibliographystyle{plain}
\bibliography{references}
\clearpage
\appendix
\section{Synthetic experiment}\label{app:synthetic}

In order to validate the findings obtained in image generation on more generic data, we test the \textit{subspace diffusion} framework on a synthetic dataset in $\mathbb{R}^{30}$. The dataset is a mixture of 100 Gaussians, each with isotropic variance $\sigma^2 = 0.05^2$, and whose centers are first sampled from a unit Gaussian and then modified such that the total variance is 50\% and 75\% explained by 6 and 11 PCA components, respectively. $64000$ samples are drawn from the mixture of Gaussians to form the training dataset and are diffused under the variance exploding SDE with $\sigma_\text{min}=0.01, \sigma_\text{max}=13$. We train simple 3-layer feedforward score models in each optimal subspace---that is, the subspaces spanned by principal components---of dimensions 1--29. With each model, we generate 6400 samples in conjunction with a full-dimensional model at varying transition times from 0--1 in increments of 0.01. We use the Euler-Maruyama solver with 100 steps and Langevin corrections with a signal-to-noise ratio of 0.2.

The sample quality is evaluated in terms of the mean $L_2$ distance from each sampled point to the nearest training point, shown below in Figure \ref{fig:synthetic}. A U-shaped trend in sample quality is again observed as the transition time is varied. All subspaces with dimension $\ge 7$ improve over the full-dimensional model (bottom row, mean distance $\approx 5.4$), reinforcing the observation made in the image generation experiments. These experiments also highlights the greater generality of subspace diffusion compared to techniques that are limited to image generation such as cascading diffusion models.

\begin{figure}[h]
    \centering
    \includegraphics[width=0.75\columnwidth]{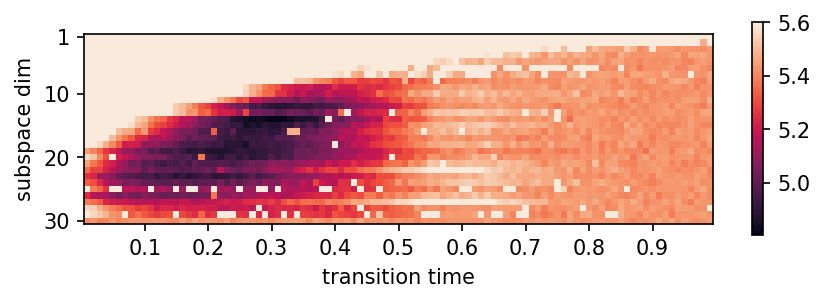}
    \caption{Results on the synthetic dataset when varying the subspace dimension and transition time. The color indicates the sample quality in terms of the mean distance from each sampled point to the nearest training point (lower / darker is better). The results from the full-dimensional model alone are shown as the subspace model of dimension 30 in the bottom row and are (as expected) constant in quality.}
    \label{fig:synthetic}
\end{figure}

\section{Patch-PCA} \label{appendix:patch-pca}
We investigate the optimality of the downsampling subspaces in comparison with the best possible subspaces that produce an image-structured latent. Recall that the downsampling subspaces are defined as follows: suppose we have a full-resolution image $\mathbf{X} \in \mathbb{R}^{(n \times n \times 3)}$, with $n$ an integer power of 2. Then the downsampled image $\mathbf{X}' \in \mathbb{R}^{(n/2 \times n/2 \times 3)}$ satisfies
\begin{equation}
    \mathbf{X}'[a, b] = \frac{1}{2}\sum_{(i,j)\in \{0,1\}^2} \mathbf{X}[2a+i, 2b+j]
\end{equation}
where each element $\mathbf{X}[i, j]$ is an RGB color in $\mathbb{R}^3$. For output pixel $\mathbf{X}'[a,b]$, this is an operation over the $2 \times 2$ patch of pixels $\mathbf{X}[2a+i,2b+j] \mid (i, j) \in \{0,1\}^2$, which can be regarded as an element of a 12-dimensional vector space. That is,
\begin{equation} \label{eq:subspace}
    \mathbf{X}'[a, b] = f(\mathbf{X}[2a+i,2b+j] \mid (i, j) \in \{0,1\}^2) \quad f: \mathbb{R}^{12} \rightarrow \mathbb{R}^3
\end{equation}
for $f$ independent of $a, b$. The downsampling subspace corresponds to taking twice the mean of the input patch, but we can generalize to arbitrary linear functions and consider \eqref{eq:subspace} with any linear $f$ to define an image-structured subspace. The key aspect of this definition is that each basis element of the subspace corresponds to a \emph{spatially localized} set of input features, and that the transformation operates identically for all spatial locations in the original image.

To find the optimal $n/2 \times n/2$ image-structured subspace we run PCA over the 12-dimensional distribution of patches $\mathbf{X}[2a+i,2b+j] \mid (i, j) \in \{0,1\}^2$ for all possible values of $(a, b)$, and over all images (or as large a subset as is computationally feasible). We then project each patch onto the top three principal components to form the smaller image. This definition and procedure can be naturally extended to smaller subspaces by considering the input patches of $4\times 4$, $8\times 8$ pixels as vector spaces of dimensionality 48, 192, etc.

\section{Hyperparameters} \label{appendix:hyperparameters}

As mentioned in the main text, we did not tune any hyperparameters for training and directly used the default settings from \cite{song2021score}, including checkpoint intervals. The sole exception was that we used reduced batch sizes due to different hardware constraints. We report FID and IS for the SDE sampler on CIFAR-10 using the best training checkpoint, as in \cite{song2021score}. All other results are obtained using the last training checkpoint.

During inference, the only hyperparameter tuned was the number of conditional Langevin steps. We tried 0, 1, 2, 5, or 10 steps using the last training checkpoint of the $8\times 8$ NCSN++ model and chose the value leading to the best FID averaged across the cutoff times. We then used 2 steps for all experiments with the SDE sampler. The Langevin signal-to-noise ratio was fixed to $0.22$ for NCSN++ and $0.01$ for DDPM++ based on the best settings found in \cite{song2021score}. All other inference hyperparameters were fixed to their default values.

\clearpage
\section{Detailed results} \label{appendix:detailed-results}

\begin{table}[h!]
    \centering
    \begin{tabularx}{\textwidth}{YYYYYYY}  
    \toprule
    Model & Subspace  & Threshold & $t_1$ & Runtime &  FID $\downarrow$ & IS $\uparrow$ \\ \midrule
    \multirowcell{13}{NCSN++\\shallow\\(VE)} & None & -- & -- & 100\% &  2.38 & 9.93\\ \cmidrule(lr){2-7}
    & \multirowcell{6}{$16 \rightarrow 32$}   
      & $1\times 10^{-4}$ & 0.64 & 75\% & 2.45 & 9.81 \\
    & & $3\times 10^{-4}$ & 0.60 & 72\% & 2.37 & 9.87 \\
    & & $1\times 10^{-3}$ & 0.56 & 69\% & 2.31 & 9.95 \\
    & & $3\times 10^{-3}$ & 0.52 & 66\% & \textbf{2.29} & \textbf{9.99} \\
    & & $1\times 10^{-2}$ & 0.47 & 63\% & 2.46 & 9.96 \\
    & & $3\times 10^{-2}$ & 0.42 & 59\% & 2.67 & 9.93 \\  \cmidrule(lr){2-7}
    & \multirowcell{6}{$8 \rightarrow 32$}   
      & $1\times 10^{-4}$ & 0.69	& 72\%	&	2.41	&	9.90 \\
    & & $3\times 10^{-4}$ & 0.64	& 68\%	&	2.39	&	9.83 \\
    & & $1\times 10^{-3}$ & 0.60	& 64\%	&	\textbf{2.29}	&	9.92 \\
    & & $3\times 10^{-3}$ & 0.56	& 60\%	&	2.35	&	10.08 \\
    & & $1\times 10^{-2}$ & 0.51	& 56\%	&	2.74	&	10.09 \\
    & & $3\times 10^{-2}$ & 0.46	& 52\%	&	3.42	&	\textbf{10.10} \\  \midrule
        \multirowcell{13}{NCSN++\\deep\\(VE)} & None & --  & -- & 100\% &  2.20 & 9.89\\ \cmidrule(lr){2-7}
    & \multirowcell{6}{$16 \rightarrow 32$}   
      & $1\times 10^{-4}$ & 0.64	& 75\%	&	2.25	&	9.86 \\
    & & $3\times 10^{-4}$ & 0.60	& 73\%	&	2.19	&	9.93\\
    & & $1\times 10^{-3}$ & 0.56	& 69\%	&	\textbf{2.17}	&	\textbf{9.94} \\
    & & $3\times 10^{-3}$ & 0.52	& 67\%	&	2.23	&	9.91 \\
    & & $1\times 10^{-2}$ & 0.47	& 63\%	&	2.31	&	9.90 \\
    & & $3\times 10^{-2}$ & 0.42	& 60\%	&	2.51	&	9.82 \\  \cmidrule(lr){2-7}
    & \multirowcell{6}{$8 \rightarrow 32$}   
      & $1\times 10^{-4}$ & 0.69	& 72\%	&	2.22	&	9.85 \\
    & & $3\times 10^{-4}$ & 0.64	& 68\%	&	2.24	&	9.87 \\
    & & $1\times 10^{-3}$ & 0.60	& 64\%	&	\textbf{2.21}	&	9.92 \\
    & & $3\times 10^{-3}$ & 0.56	& 60\%	&	2.39	&	\textbf{10.08} \\
    & & $1\times 10^{-2}$ & 0.51	& 56\%	&	3.05	&	10.01 \\
    & & $3\times 10^{-2}$ & 0.46	& 51\%	&	3.51	&	9.96 \\  \bottomrule
    \end{tabularx}
    \caption{NCSN++ subspace diffusion results on CIFAR-10 unconditional generation. Runtimes are reported as percentages of the respective full diffusion model.}
    \label{tab:cifar_ncsn}
\end{table}

\begin{table}[h!]
    \centering
    \begin{tabularx}{\textwidth}{YYYYYYY}  
    \toprule
    Model & Subspace  & Threshold & $t_1$ & Runtime &  FID $\downarrow$ & IS $\uparrow$ \\ \midrule
    \multirowcell{13}{DDPM++\\shallow\\(sub-VP)} & None  & -- & -- & 100\% &  2.61 & 9.56\\ \cmidrule(lr){2-7}
    & \multirowcell{6}{$16 \rightarrow 32$}   
      & $1\times 10^{-4}$ &	0.56 & 69\%	&	\textbf{2.61}	&	9.53 \\
    & & $3\times 10^{-4}$ &	0.51 & 65\%	&	2.63	&	9.64 \\
    & & $1\times 10^{-3}$ &	0.45 & 61\%	&	2.75	&	\textbf{9.66} \\
    & & $3\times 10^{-3}$ &	0.39 & 56\%	&	3.11	&	9.53 \\
    & & $1\times 10^{-2}$ &	0.32 & 52\%	&	4.07	&	9.52 \\
    & & $3\times 10^{-2}$ &	0.26 & 47\%	&	5.68	&	9.37 \\  \cmidrule(lr){2-7}
    & \multirowcell{6}{$8 \rightarrow 32$}   
      & $1\times 10^{-4}$ &	0.62 & 65\%	&	\textbf{2.60}	&	9.54 \\
    & & $3\times 10^{-4}$ &	0.57 & 60\%	&	2.68	&	9.56 \\
    & & $1\times 10^{-3}$ &	0.50 & 55\%	&	2.93	&	\textbf{9.66} \\
    & & $3\times 10^{-3}$ &	0.45 & 50\%	&	3.73	&	9.63 \\
    & & $1\times 10^{-2}$ &	0.38 & 43\%	&	5.24	&	9.51 \\
    & & $3\times 10^{-2}$ &	0.31 & 37\%	&	7.61	&	9.23 \\  \midrule
        \multirowcell{13}{DDPM++\\deep\\(sub-VP)} & None & -- & -- & 100\% &  2.41 & 9.57\\ \cmidrule(lr){2-7}
    & \multirowcell{6}{$16 \rightarrow 32$}   
      & $1\times 10^{-4}$ &	0.56 & 69\%	&	\textbf{2.40}	&	9.66 \\
    & & $3\times 10^{-4}$ &	0.50 & 66\%	&	2.43	&	9.62 \\
    & & $1\times 10^{-3}$ &	0.44 & 61\%	&	2.55	&	9.65 \\
    & & $3\times 10^{-3}$ &	0.38 & 57\%	&	2.84	&	\textbf{9.68} \\
    & & $1\times 10^{-2}$ &	0.32 & 53\%	&	3.49	&	9.55 \\
    & & $3\times 10^{-2}$ &	0.26 & 48\%	&	4.64	&	9.52 \\  \cmidrule(lr){2-7}
    & \multirowcell{6}{$8 \rightarrow 32$}   
      & $1\times 10^{-4}$ &	0.62 & 65\%	&	\textbf{2.46}	&	9.67 \\
    & & $3\times 10^{-4}$ &	0.56 & 60\%	&	2.52	&	9.67 \\
    & & $1\times 10^{-3}$ &	0.50 & 55\%	&	2.76	&	\textbf{9.72} \\
    & & $3\times 10^{-3}$ &	0.44 & 50\%	&	3.41	&	9.65 \\
    & & $1\times 10^{-2}$ &	0.38 & 43\%	&	4.39	&	9.55 \\
    & & $3\times 10^{-2}$ &	0.31 & 37\%	&	6.32	&	9.30 \\  \bottomrule
    \end{tabularx}
    \caption{DDPM++ subspace diffusion results on CIFAR-10 unconditional generation. Runtimes are reported as percentages of the respective full diffusion model.}
    \label{tab:cifar_ddpm}
\end{table}

\clearpage
\section{Additional samples} \label{appendix:samples}

\begin{figure}[h!]
    \centering
    \includegraphics[width=\textwidth]{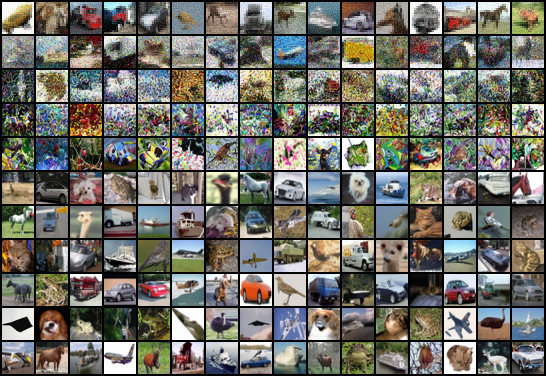}
    \caption{Random samples from CIFAR-10 using the NCSN++ deep $16\times 16$ subspace diffusion. Each row shows samples with an extra 10\% of the diffusion on the  full-dimensional space (from 0\% at the top to 100\% at the bottom). High quality samples start appearing with 50--60\% of the diffusion at full dimensionality.  }
    \label{fig:cifar16}
\end{figure}

\begin{figure}
    \centering
    \includegraphics[width=\textwidth]{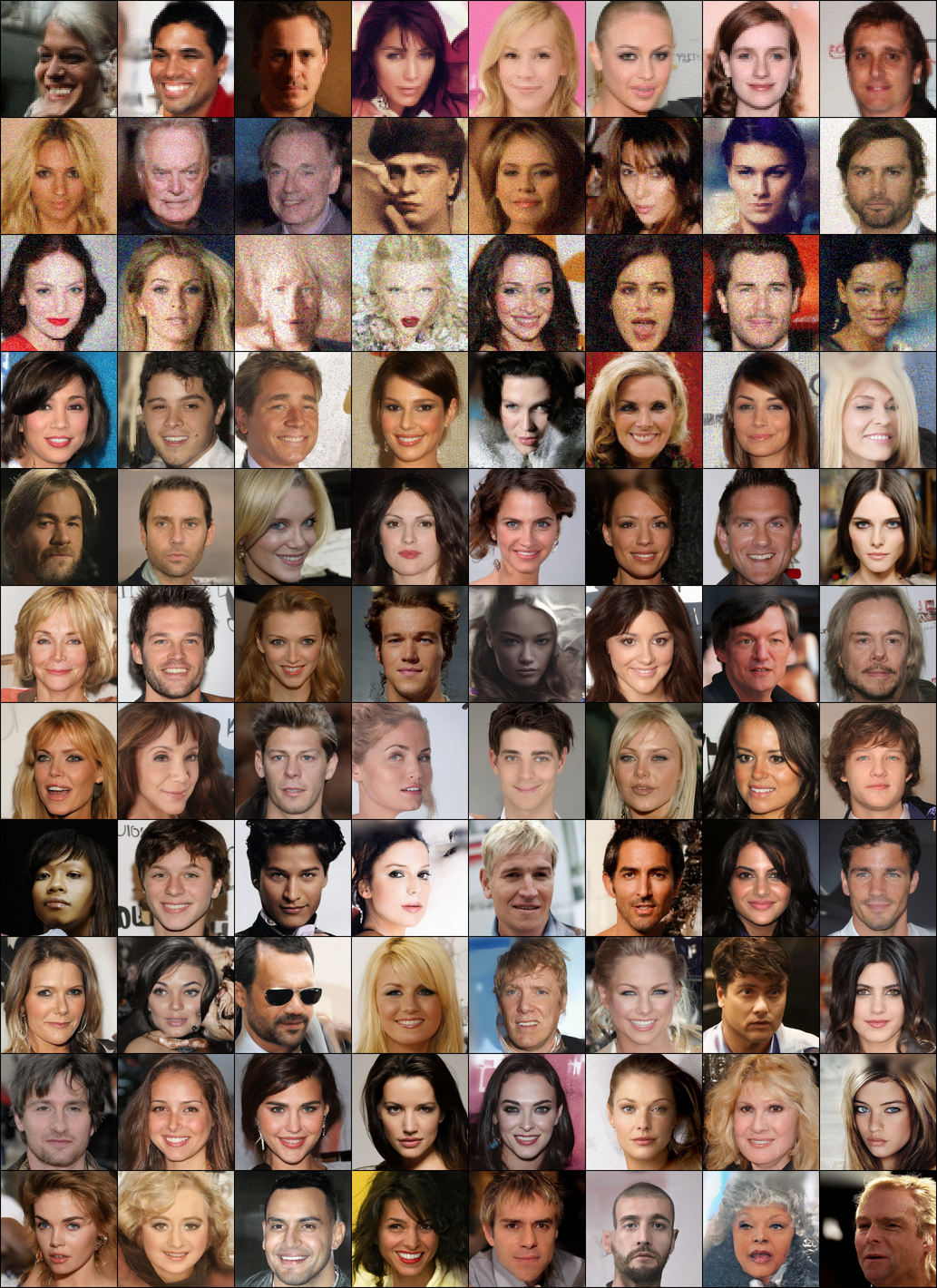}
    \caption{Random samples from CelebA-HQ using the NCSN++ $64\times 64$ subspace diffusion. Each row shows samples with an extra 10\% of the diffusion on the  full-dimensional space (from 0\% at the top to 100\% at the bottom). High quality samples start appearing with 30-40\% of the diffusion at full dimensionality.  }
    \label{fig:celeba64}
\end{figure}

\begin{figure}
    \centering
    \includegraphics[width=\textwidth]{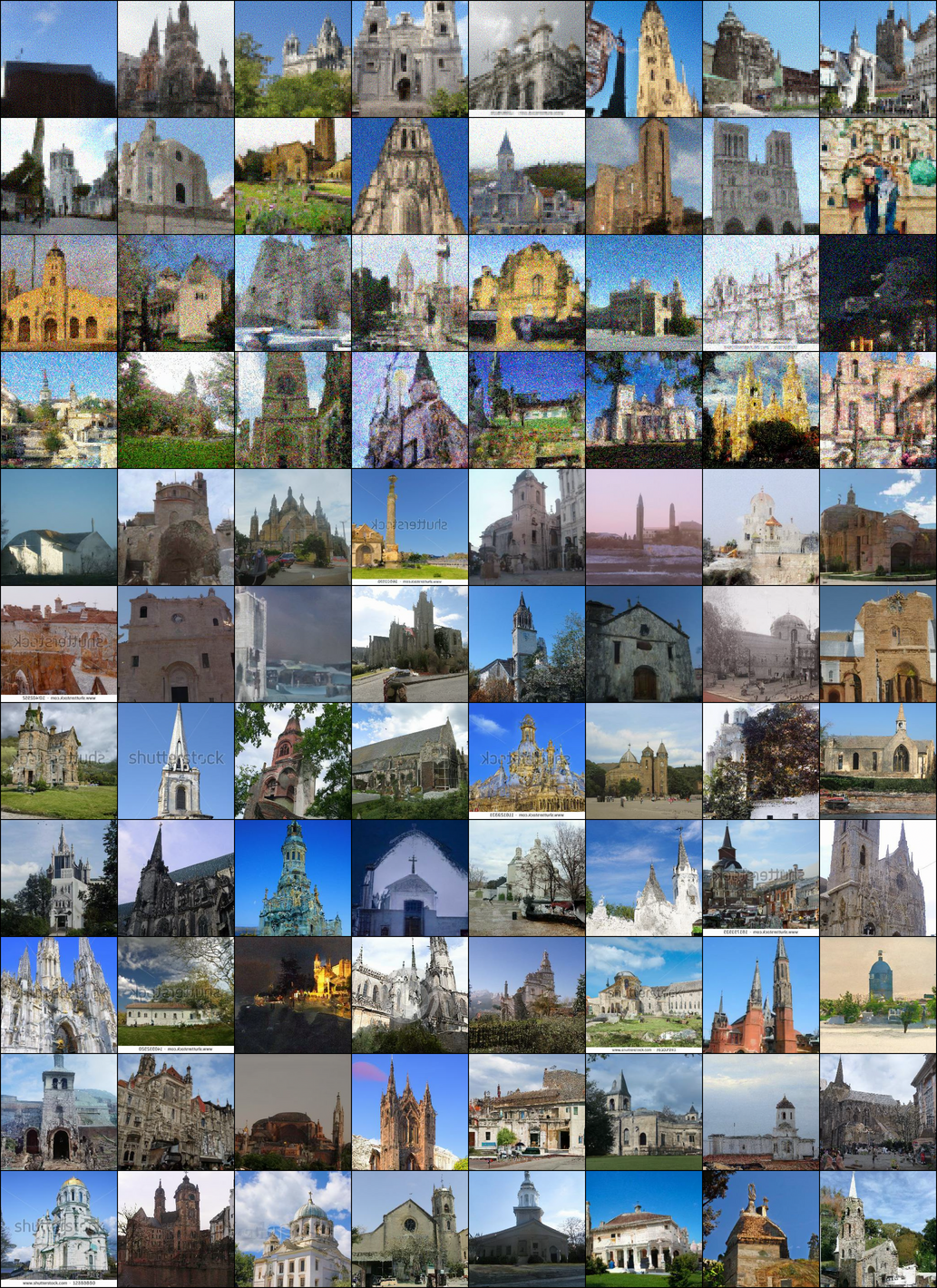}
    \caption{Random samples from LSUN Church using the NCSN++ $64\times 64$ subspace diffusion. Each row shows samples with an extra 10\% of the diffusion on the  full-dimensional space (from 0\% at the top to 100\% at the bottom). High quality samples start appearing with 40\% of the diffusion at full dimensionality.}
    \label{fig:church64}
\end{figure}

\begin{figure}
    \centering
    \includegraphics[width=\textwidth]{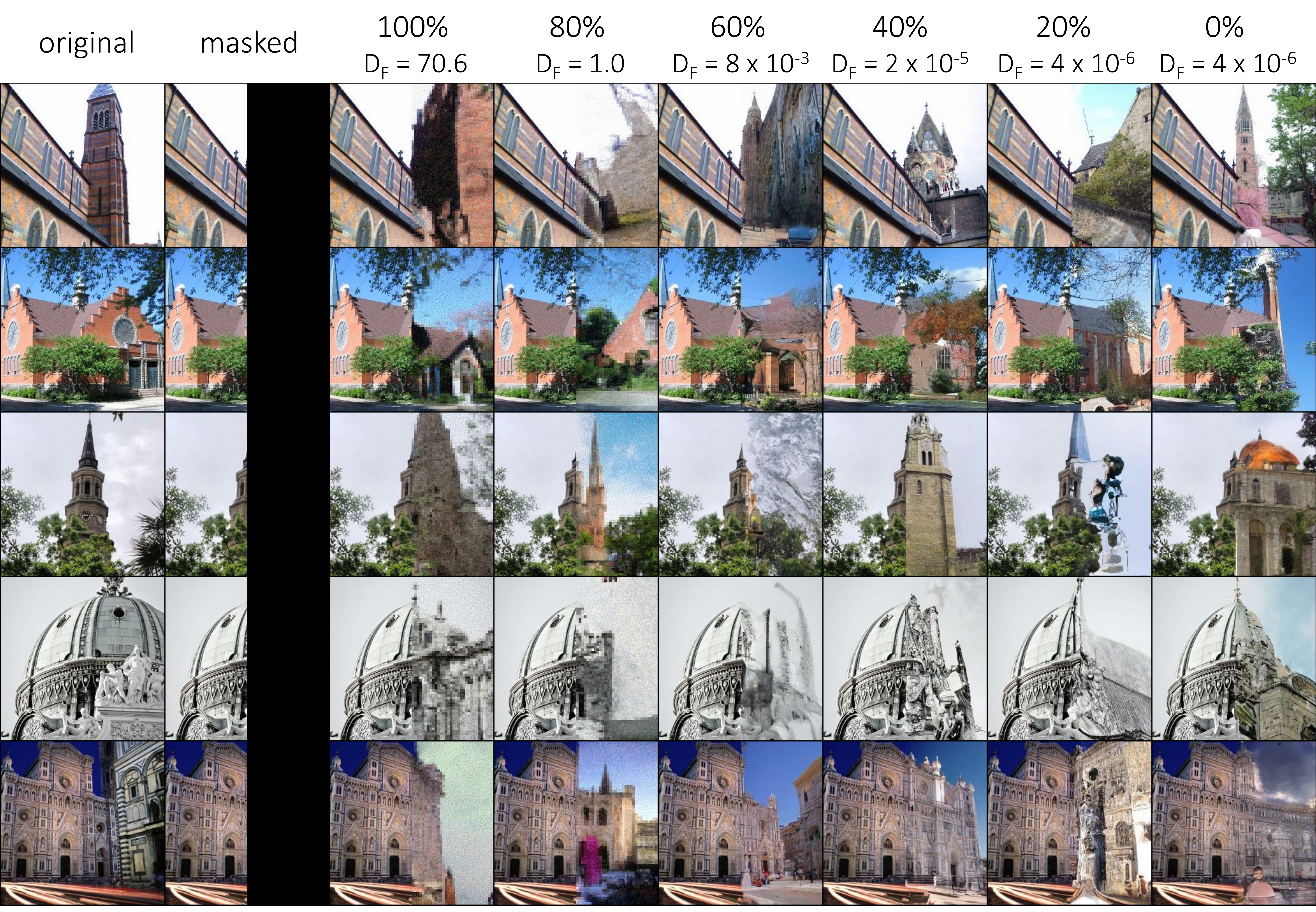}
    \caption{Random samples of inpainting procedure from LSUN Church using the NCSN++ $64\times 64$ subspace diffusion, with different proportions of subspace diffusion (reported at the top along with the corresponding Fisher divergence threshold).  }
    \label{fig:inpainting}
\end{figure}
\end{document}